\documentclass[10pt,twocolumn,letterpaper]{article}

\usepackage{cvpr}              
\usepackage{esvect}
\usepackage{stfloats}
\usepackage{multirow}

\usepackage{threeparttable} 

\usepackage[accsupp]{axessibility}

\definecolor{cvprblue}{rgb}{0.21,0.49,0.74}
\usepackage[pagebackref,breaklinks,colorlinks,citecolor=cvprblue]{hyperref}

\title{Prior Does Matter: Visual Navigation via Denoising Diffusion Bridge Models}

\author{Hao Ren\textsuperscript{1}, Yiming Zeng\textsuperscript{1}, Zetong Bi\textsuperscript{1}, Zhaoliang Wan\textsuperscript{1}, Junlong Huang\textsuperscript{2}, Hui Cheng\textsuperscript{1}\thanks{Corresponding author: Hui Cheng (chengh9@mail.sysu.edu.cn).}\\
\textsuperscript{1}School of Computer Science and Engineering, Sun Yat-sen University\\
\textsuperscript{2}School of Intelligent Systems Engineering, Sun Yat-sen University
}

\begin{document}
\maketitle
\begin{abstract}

Recent advancements in diffusion-based imitation learning, which shows impressive performance in modeling multimodal distributions and training stability, have led to substantial progress in various robot learning tasks. In visual navigation, previous diffusion-based policies typically generate action sequences by initiating from denoising Gaussian noise.
However, the target action distribution often diverges significantly from Gaussian noise, leading to redundant denoising steps and increased learning complexity. Additionally, the sparsity of effective action distributions makes it challenging for the policy to generate accurate actions without guidance.
To address these issues, we propose a novel, unified visual navigation framework leveraging the denoising diffusion bridge models named NaviBridger. This approach enables action generation by initiating from any informative prior actions, enhancing guidance and efficiency in the denoising process.
We explore how diffusion bridges can enhance imitation learning in visual navigation tasks and further examine three source policies for generating prior actions. Extensive experiments in both simulated and real-world indoor and outdoor scenarios demonstrate that NaviBridger accelerates policy inference and outperforms the baselines in generating target action sequences.
Code is available at \url{https://github.com/hren20/NaiviBridger}.

\end{abstract}    
\section{Introduction}
\label{sec:introduction}

\begin{figure}[htbp]
    \centering
    \includegraphics[width=\linewidth]{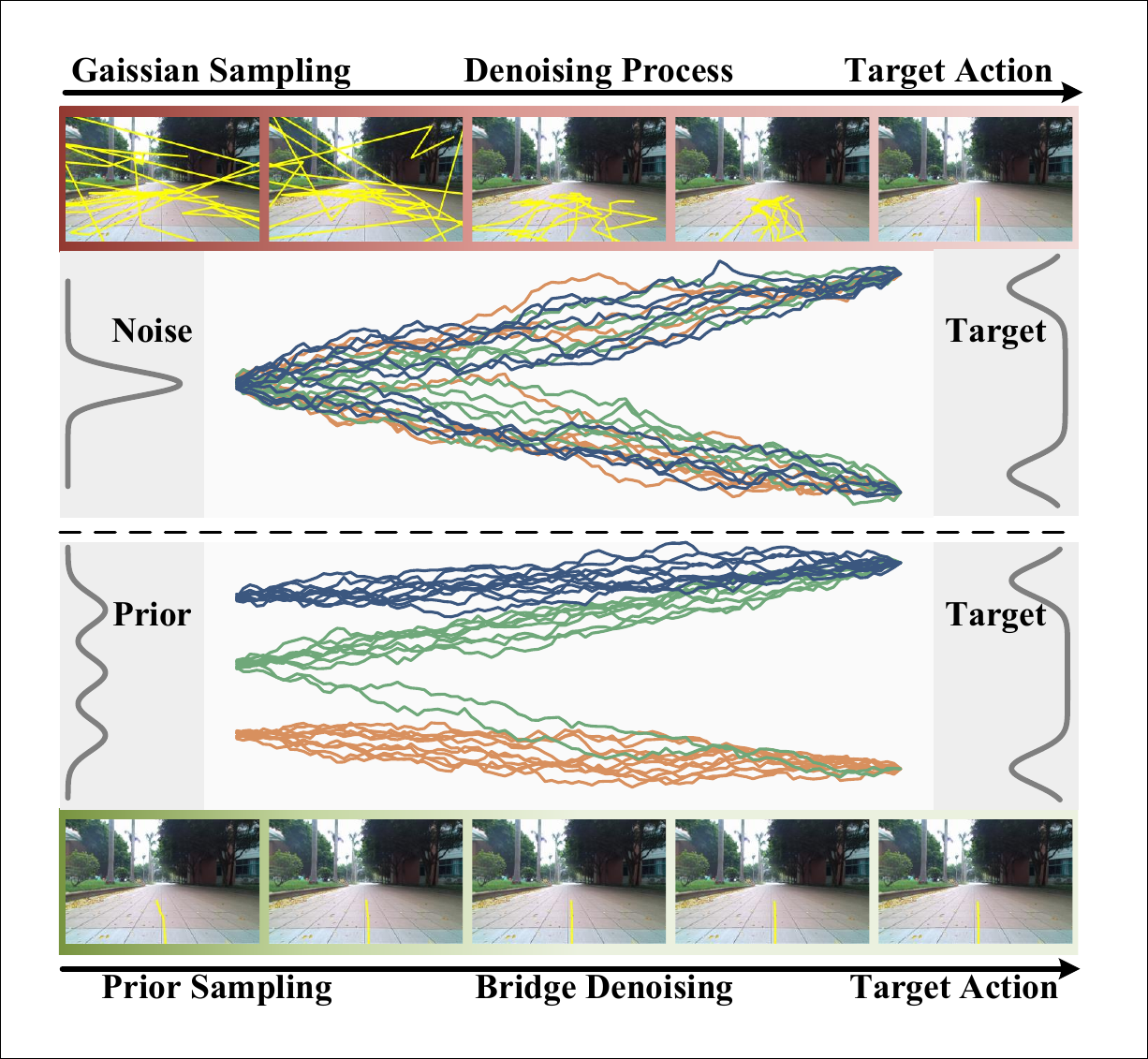}
    \caption{Visual representation overview of the local path generation process. The top row shows the actions generated through vanilla diffusion models denoising from Gaussian noise. The bottom row illustrates actions generated using proposed NaviBridger, which leverage prior information to effectively guide the denoising bridge model towards the target action. This comparison demonstrates the importance of incorporating prior knowledge in achieving stable and accurate navigation.}
    \label{fig:teaser}
    \vspace{-16pt}
\end{figure}

Visual navigation \cite{bonin2008visual, zhang2022survey, zhu2017target, shah2023gnm} is vital for mobile robots, allowing them to reach targets like images or locations using visual observation. 
Previous methods typically consist of a global planner that provides high-level instructions or subgoals, often leveraging large language models \cite{chen2021topological, park2023visual, shah2023lm} or topological maps \cite{sridhar2024nomad}, to guide a local planner that generates feasible local paths in real time.
Imitation learning from expert demonstrations for local path planning \cite{wu2020towards, shah2023vint} is emerging as a promising research direction, with advantages in modeling multimodal distributions and capturing sequence correlation.

Despite the significant progress of generative model-based local planners, this field faces two major issues due to the characteristics of generative models. First, substantial distributional discrepancies between the Gaussian noise, typically used in these models, and the robot's target action distribution lead to redundant denoising steps during inference. Second, the sparsity of desired actions makes it challenging for the policy to generate accurate actions, especially when denoising starts from chaotic random noise. Together, these issues increase computational costs and degrade overall performance.

As Oscar Wilde stated, "Without order, nothing can exist; without chaos, nothing can evolve," prompting an intriguing question: Could a structured, task-specific distribution serve as the starting point for denoising in diffusion models rather than relying on chaotic Gaussian noise? This idea aligns with optimal transport theory, which addresses translations between arbitrary distributions \cite{chen2016entropic, villani2009optimal}, and has been successfully applied in diffusion models for paired-data tasks like image translation \cite{li2023bbdm} and restoration \cite{yue2024enhanced}. However, diffusion bridge methods cannot be directly applied in robot learning tasks such as visual navigation, where no inherent source distribution exists to derive the target distribution. 
Therefore, the key challenge lies in constructing a suitable initial distribution and adapting the diffusion bridge techniques specifically for visual navigation.

To address this, we introduce \textbf{NaviBridger}, a novel, unified visual navigation framework that leverages Denoising Diffusion Bridge Models (DDBM) \cite{zhou2024denoising}. NaviBridger redefines action generation by initiating from informative prior actions rather than random Gaussian noise, enhancing guidance and efficiency in the denoising process. We analyze its application in imitation learning, presenting key computational formulas, and provide a theoretical analysis revealing a crucial insight: the quality of the target action distribution improves when derived from a more suitable source distribution. Based on this insight, We introduce various initialization methods tailored to different source distribution types, including uninformative noise priors, informative handcrafted priors, and learning-based priors, with a detailed discussion of the advantages and limitations. NaviBridger also unifies previous visual navigation methods that relied on traditional diffusion models, and maintains compatibility with Gaussian noise when no suitable source distribution prior is available. NaviBridger is, to our knowledge, the first work to explore diffusion bridge models for visual navigation. Our contributions are as follows:

\begin{itemize}
    \item[$\bullet$] \textbf{General Framework for Visual Navigation}:
    We introduce NaviBridger, a visual navigation framework based on the denoising diffusion bridge model, which uses prior actions instead of Gaussian noise as the initial input, achieving an effective balance between performance and computational efficiency. This framework can be easily adapted to a variety of imitation learning tasks.
    \item[$\bullet$] \textbf{Theoretical Analysis of Source Distributions}: We theoretically analyze the impact of source distributions on diffusion-based model performance, presenting tailored methods for generating prior source actions, including Gaussian, rule-based, and learning-based priors.
    \item[$\bullet$] \textbf{Comprehensive Evaluation of NaviBridger}: We conducted a comprehensive evaluation of our method across various scenarios and tasks, examining the effectiveness of the framework and the key design decisions of various source distributions and denoising steps. Experimental results show that NaviBridger exceeds baseline models across multiple scenarios.
    
\end{itemize}
\section{Related Work}
\label{sec:related}

\textbf{Diffusion  Models} \cite{sohl2015deep, ho2020denoising} have rapidly advanced in areas such as conditional generation \cite{dhariwal2021diffusion, ho2022classifier, zeng2024lvdiffusor}, faster training \cite{song2020denoising}, and theoretical refinements \cite{song2020score, karras2022elucidating}. Their strong generative capabilities are widely applied in image restoration \cite{lugmayr2022repaint}, image translation \cite{saharia2022palette}, and text-to-image generation \cite{rombach2022high}. Recently, diffusion models have shown potential in decision-making fields like reinforcement learning \cite{zhu2023diffusion} and imitation learning \cite{pmlrv162janner22a, ajay2022conditional}.

Diffusion bridges, initially a tool in probability theory, have evolved into a key approach in generative modeling \cite{liu2022let, somnath2023aligned}. \cite{sarkka2019applied} laid the groundwork, and \cite{heng2021simulating} explored diffusion bridges conditioned on fixed start and endpoints. Recent work extends these ideas by applying Doob's $h$-transform for bridging arbitrary distributions \cite{liu2022let}. \cite{li2023bbdm} proposes reversing Brownian Bridges for distribution translation in discrete time, while \cite{zhou2024denoising} constructs bridge models using VP and VE diffusion processes in continuous time.

\textbf{Visual Navigation}. 
Traditional navigation approaches \cite{borenstein1989real, borenstein1991vector, oriolo1995line} typically divide the navigation task into three main components: perception and mapping \cite{cadena2016past, ren2023adaptive, ren2025layer}, localization \cite{yasuda2020autonomous, chalvatzaras2022survey}, and path planning \cite{yang2016survey}, with each element showing considerable advancements in recent studies. However, the modular design can lead to cumulative information loss and rely heavily on precise measurements. Since Zhu et al. \cite{zhu2017target} pioneered the training of agents in simulated environments, learning-based visual navigation has progressed rapidly \cite{al2022zero, beeching2020learning, chen2021topological, du2021curious, majumdar2022zson}. \cite{zhu2017target} employed deep reinforcement learning to address visual navigation challenges for image-based target search in small indoor simulation scenarios. Subsequent research has focused on improving visual representation by incorporating supplementary information\cite{hong2023learning}, auxiliary tasks\cite{roth2024viplanner}, or optimized networks \cite{zeng2021pushing}. By contrast, recent works such as \cite{loquercio2018dronet, hirose2023exaug, shah2023gnm} emphasize navigation tasks in real-world robotic systems, aiming to achieve zero-shot or few-shot navigation. ViNT \cite{shah2023vint} introduced a foundational model based on topological mapping for long-range navigation tasks and implemented the generation of latent subgoal images using diffusion models.

Most recently, state-conditioned diffusion models for imitation learning in planning are particularly powerful\cite{pmlrv162janner22a, urain2023se, chi2023diffusion, zeng2025navidiffusor}. NoMaD \cite{sridhar2024nomad} incorporates diffusion models into visual navigation, enabling the direct generation of actions and achieving robust performance and enhanced multimodal representations. However, these approaches rely on standard diffusion models that denoise progressively from an initial Gaussian noise distribution. For more complex action spaces, this "sampling from an initial Gaussian state" approach often demands additional denoising steps \cite{chen2024behavioral}. Furthermore, since the Gaussian distribution differs significantly from the target action distribution, fitting the model to the target distribution becomes more challenging, potentially reducing the quality of generated actions \cite{de2021diffusion, park2024leveraging}.
\section{Preliminaries}
\label{sec:preliminary}

\begin{figure*}[htbp]
    \centering
    \includegraphics[width=\linewidth]{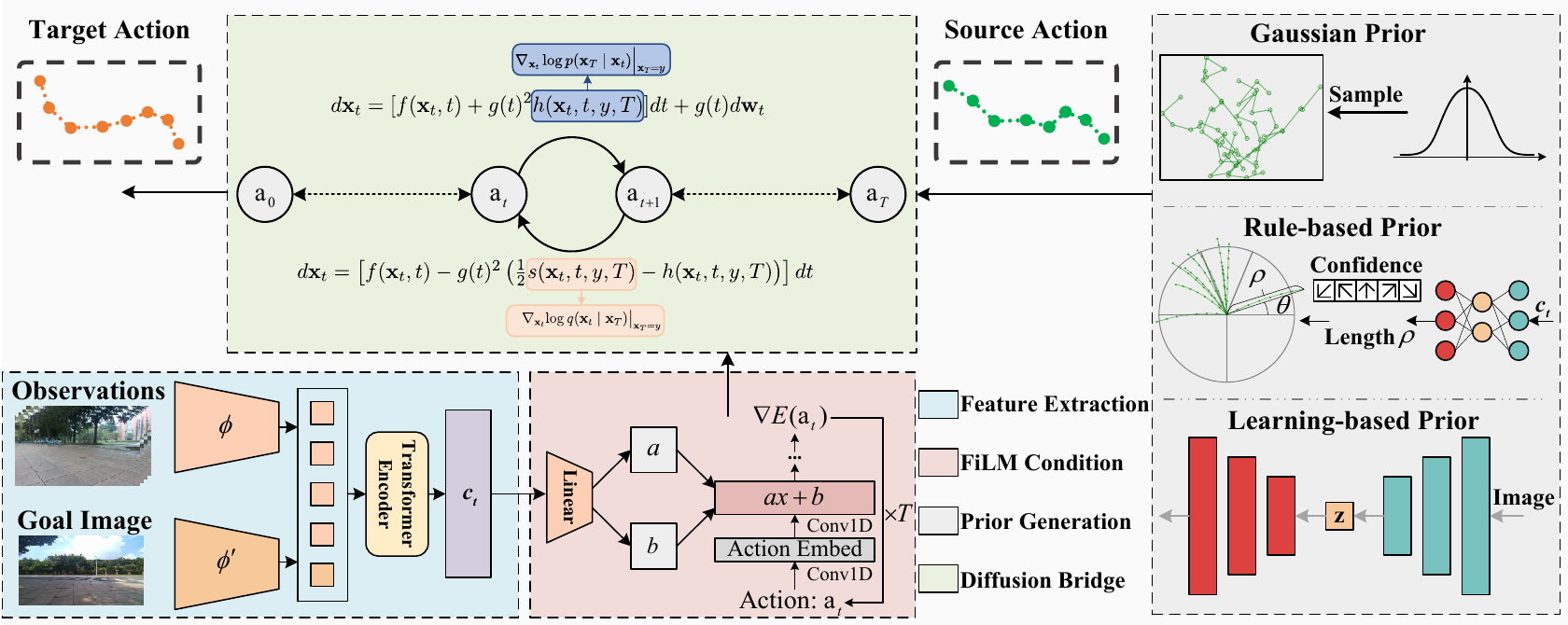}
    \caption{\textbf{NaviBridger Overview}. The policy takes RGB observations and a goal image, extracting features via a Transformer encoder. 
    The FiLM condition module applies learned conditions to improve path accuracy, while the prior generation module ensures the policy outputs better aligns with target actions, starting from alternative prior actions generated using optional prior policies.}
    \label{fig:main}
\end{figure*}

\textbf{Diffusion Process}: A diffusion process gradually adds noise to data, translating it from an initial distribution \( p_0 \) to a simpler, typically Gaussian distribution \( p_T = \mathcal{N}(0, I) \). This forward process is modeled by the stochastic differential equation (SDE):

\begin{equation}
d\mathbf{x}_t = f(\mathbf{x}_t, t) dt + g(t) d\mathbf{w}_t,
\label{eq:1}
\end{equation}
where \( f \) is the drift function, \( g(t) \) is the diffusion coefficient, and \( \mathbf{w}_t \) represents the Wiener process.

The reverse process, which removes noise to reconstruct data, is governed by the reverse-time SDE:

\begin{equation}
d\mathbf{x}_t = [f(\mathbf{x}_t, t) - g(t)^2 \nabla_{\mathbf{x}_t} \log p(\mathbf{x}_t)] dt + g(t) d\bar{\mathbf{w}}_t,
\end{equation}
where \( \nabla_{\mathbf{x}_t} \log p(\mathbf{x}_t) \) is the score function.

Diffusion  Models generate images by progressively adding noise to data and then training neural networks to reverse this process, removing noise to recreate the original image. However, diffusion models are limited because they can only transport complex data distributions to a standard Gaussian distribution. Building upon the study of stochastic differential equations, using a diffusion process to map data to a known fixed endpoint is achievable through Doob's $h$-transform.

\textbf{Diffusion Bridge}: A diffusion bridge extends the diffusion process by interpolating between two fixed distributions, ensuring the generated data transitions from a starting distribution to a specified endpoint. The forward process which is expanded from Eq.(\ref{eq:1}) of diffusion bridge is expressed as:

\begin{equation}
d\mathbf{x}_t = [f(\mathbf{x}_t, t) + g(t)^2 h(\mathbf{x}_t, t, y, T)] dt + g(t) d\mathbf{w}_t,
\end{equation}
where $\boldsymbol{h}(\mathbf{x}_t, t, y, T) = \nabla_{\mathbf{x}_t}\log p(\mathbf{x}_T\mid \mathbf{x}_t)\big \rvert_{\mathbf{x}_t = x, \mathbf{x}_T=y}$ is the Doob's $h$-transform ensures approximation to the target distribution \( y \) at time \( T \).

\textbf{Denoising Diffusion Bridge Models (DDBM)}: DDBMs generalize diffusion models by enabling transport between paired distributions, using reverse-time SDE or probability flow ODE. For the reverse diffusion bridge, the SDE is:

\begin{equation}
\begin{split}
d\mathbf{x}_t &= [f(\mathbf{x}_t, t) - g(t)^2 (s(\mathbf{x}_t, t, y, T) \\
            &\quad - h(\mathbf{x}_t, t, y, T))] dt + g(t) d\bar{\mathbf{w}}_t,
\end{split}
\end{equation}
where $\mathbf{s}(\mathbf{x}_t, t, y, T) = \nabla_{\mathbf{x}_t}{\log q(\mathbf{x}_t\mid \mathbf{x}_T)}\big \vert_{\mathbf{x}_t = x, \mathbf{x}_T = y}$ is the score function, \( h(\mathbf{x}_t, t, y, T) \) ensures the transition to the target distribution. DDBM employs the sampling distribution $q(x_t \mid x_0, x_T)$ as an approximation for the challenging-to-obtain distribution $q(\mathbf{x}_t \mid \mathbf{x}_T)$, offering a defined expression for $q(x_t \mid x_0, x_T)$ and reparameterizing it to facilitate denoising (details provided in the Appendix). DDBMs use denoising score matching to learn the score function:
\begin{equation}
L_\theta\! =\! \mathbb{E}_{\mathbf{x}_t, \mathbf{x}_0, t}\!\left[w(t)\|s_\theta(\mathbf{x}_t,\! \mathbf{x}_T,\! t)\! -\! \nabla_{\mathbf{x}_t}\! \log q(\mathbf{x}_t | \mathbf{x}_0,\! \mathbf{x}_T)\|^2\right]
\end{equation}
where $w(t)$ indicates the weight scheduler of the loss.

This method allows DDBMs to translate between distributions effectively, expanding the scope of diffusion models to a wide range of generative tasks. More details about DDBM refer to the Appendix.
\section{Methodology}
\label{sec:method}

Motivated by the preceding discussion, we propose a novel approach, Navibridger. Instead of relying on sampling and denoising within a fixed Gaussian noise space, our method transforms this process into a denoising procedure that maps from an observation-based arbitrary source distribution policy $\pi_s$ to a target distribution policy $\pi$.
Navibridger consists of three core components: a feature extraction module, a prior action generation module, and a denoising diffusion bridge module. The architecture illustration is shown in Fig. \ref{fig:main}.

In visual navigation, the objective is to design a control policy $\pi$ that allows a robot to execute target-oriented navigation effectively. The inputs to this policy consist of the robot's current and past visual observation sequence, denoted by $\mathcal{O} = \{\boldsymbol{I}_t\}^T_{t=T-p}$, along with a future action sequence $\mathbf{a}=\{W_t\}^{T+n}_{t=T}$. Here, $p$ represents the past time window. The policy can also incorporate a goal image $\boldsymbol{I}_g$, specifying the navigation target. Each observation $\boldsymbol{I}_i$ is mapped to a feature vector $\phi(\boldsymbol{I}_i)$, which is subsequently processed by a Transformer encoder to produce a contextual vector $\boldsymbol{c}_t$ in the usual way. The context vector $\boldsymbol{c}_t$ is input into the prior action generation module to produce the source action $\mathbf{a}_{s}\in \pi_s$ associated with $\boldsymbol{I}_t$. The diffusion bridge model translates this source action into the desired target action $\mathbf{a}\in \pi$.

\subsection{Diffusion Bridge for Imitation Learning}
\label{subsec:ddbmil}
In the imitation learning framework based on DDBM, the model employs a reverse diffusion process to progressively adjust the initial action distribution continuously, resulting in an action trajectory that satisfies the target requirements. Specifically, the DDBM reverse diffusion process for imitation learning can be described as follows:
\begin{equation}
\begin{split}
d \mathbf{a}_t = &[f(\mathbf{a}_t, t) - g^2(t) (s(\mathbf{a}_t, t, \mathbf{a}_T, T)\\
&- h(\mathbf{a}_t, t, \mathbf{a}_T, T))] dt + g(t) d\hat{w}_t,
\end{split}
\end{equation}
where \( f(\mathbf{a}_t, t) \) denotes the drift term, \( g(t) \) is the diffusion coefficient, \( s(\mathbf{a}_t, t, \mathbf{a}_T, T) \) represents the score function that gives the conditional score at time \( t \) for the current state \( \mathbf{a}_t \) relative to the target state \( \mathbf{a}_0 \), and \( h(\mathbf{a}_t, t, \mathbf{a}_T, T) \) is the Doob's \( h \)-transform, applied to adjust the diffusion path, thereby ensuring a controlled transition from \( \mathbf{a}_s \) to \( \mathbf{a}_T \).

\( h(\mathbf{a}_t, t, \mathbf{a}_T, T) \) is a function of $\nabla_{\mathbf{a}_t}\log p(\mathbf{a}_T\mid \mathbf{a}_t)$ , which is approximated by $q(\mathbf{a}_t \mid \mathbf{a}_0, \mathbf{a}_T)$:
\begin{equation}\label{eq:marginal}
    q(\mathbf{a}_t\mid \mathbf{a}_0, \mathbf{a}_T) = \mathcal{N}(\hat{\mu}_t, \hat{\sigma}_t^2\boldsymbol{I}),\quad
\end{equation}

\begin{equation}
\hat{\mu}_t= \frac{\text{SNR}_T}{\text{SNR}_t} \frac{\alpha_t}{\alpha_T}\mathbf{a}_T + \alpha_t \mathbf{a}_0(1-\frac{\text{SNR}_T}{\text{SNR}_t}),
\end{equation}

\begin{equation}
\hat{\sigma}_t^2=\sigma_t^2(1-\frac{\text{SNR}_T}{\text{SNR}_t}),
\end{equation}
where $\alpha_t$ and $\sigma_t$ denote predefined signal and noise schedules, with the signal-to-noise ratio at time $t$ given by $\normalfont\text{SNR}_t = \alpha_t^2/\sigma_t^2$.
With the reparametrization of Elucidating Diffusion Models, score function $\mathbf{s}(\mathbf{a},t,\mathbf{a}_T,T)$ can be expressed as:

\begin{equation}
\label{eq:reparam}
\begin{aligned}
  &\nabla_{\mathbf{a}_t} \text{log}q(\mathbf{a}_t|\mathbf{a}_T) \approx \mathbf{s}({D_\theta}, \mathbf{a}_t, t, \mathbf{a}_T, T) \\ &= \frac{\mathbf{a}_t - (\frac{SNR_T}{SNR_t}\frac{\alpha_t}{\alpha_T}\mathbf{a}_T + \alpha_t D_{\theta}(\mathbf{a}_t,t, \mathbf{a}_T)(1-\frac{SNR_T}{SNR_t}))}{\sigma^2_t(1-\frac{SNR_T}{SNR_t})},
\end{aligned}
\end{equation}
here, ${D_\theta}(\mathbf{a}_t,t,\mathbf{a}_T)$ denotes the output of the network.

Based on the above inferences, we developed a foundational imitation learning framework utilizing diffusion bridges. This approach theoretically enables action sampling by transferring samples $\mathbf{a}_T$ from the source distribution $\pi_s$ to the target distribution $\pi$, with neural networks employed to learn the reverse denoising process.

\subsection{Prior Policy Design}
\label{subsec:priordesign}
Previous research on diffusion bridge models has primarily focused on tasks like image translation and style transfer, which involve paired input-output tasks where the input corresponds to a source distribution sample. 
Compared to Gaussian noise, a well-defined source distribution offers an informative prior, often yielding more accurate outcomes. 

However, in imitation learning tasks, such as visual navigation, the target action is not inherently paired with an initial action. 
In this section, we present the theoretical foundation for designing the source distribution and introduce three distinct strategies for generating the source distribution specifically for visual navigation tasks.

\textbf{Error Bound Analyzing}: 
For a diffusion process to be effective, the initial source distribution must resemble the target distribution. If there is a significant discrepancy between the two, the diffusion process will require more extensive adjustments per step, amplifying drift, and score correction, thus increasing the probability of error in the likelihood flow ODE or SDE. 

Consider the error of the model denoising process. \( \mathbb{E}_{\mathbf{a}_0} [\| \mathbf{a}_t - \mathbf{a}_0 \|^2] \). This error is driven by the difference between the target distribution \( \pi(\mathbf{a}_0) \) and the source distribution \( \pi_{s}(\mathbf{a}_T) \), which can be represented using the Kullback-Leibler (KL) divergence as:
\begin{equation}
D_{\text{KL}}\left(\pi_{s}(\mathbf{a}_T) \| \pi(\mathbf{a}_0)\right) = \int \pi_{s}(\mathbf{a}_T) \log \frac{\pi_{s}(\mathbf{a}_T)}{\pi(\mathbf{a}_0)} d\mathbf{a}_T.
\end{equation}

This KL divergence measures how much information is "lost" if we approximate \( \pi(\mathbf{a}_0) \) by \( \pi_{s}(\mathbf{a}_T) \). A more minor divergence indicates a smaller amount of drift, and noise adjustment will be necessary for the DDBM to match the target.
A probability flow ODE controls the evolution of samples in a DDBM:
\begin{equation}
\frac{d\mathbf{a}_t}{dt} = f(\mathbf{a}_t, t) - g^2(t) \left(\frac{1}{2} \mathbf{s}(\mathbf{a}_t,t,\mathbf{a}_T,T) - h(\mathbf{a}_t, t, \mathbf{a}_T, T)\right)
\end{equation}
where \( \mathbf{s}(\mathbf{a}_t,t,\mathbf{a}_T,T) - h(\mathbf{a}_t, t, \mathbf{a}_T, T) \) represents the adjustment needed to shift \( \mathbf{a}_t \) closer to \( \mathbf{a}_0 \). The error bound for the model’s convergence to \( \mathbf{a}_T \) can be formulated as:
\begin{equation}
\mathbb{E}_{\mathbf{a}_t, \mathbf{a}_0} \left[ \| \mathbf{a}_t - \mathbf{a}_0 \|^2 \right] \leq C \cdot D_{\text{KL}}\left(\pi_{s}(\mathbf{a}_T) \| \pi(\mathbf{a}_0)\right),
\end{equation}
where \( C \) is a constant that depends on factors such as the noise schedule \( g(t) \) and the time duration \( T \). Hence, the error bound shows that the closer the source distribution \( q_{\text{source}} \) is to the target distribution \( q_{\text{target}} \), the lower the error bound for the DDBM, resulting in a more accurate diffusion process and improved model performance.

\begin{figure}[htbp]
    \centering
    \includegraphics[width=\linewidth]{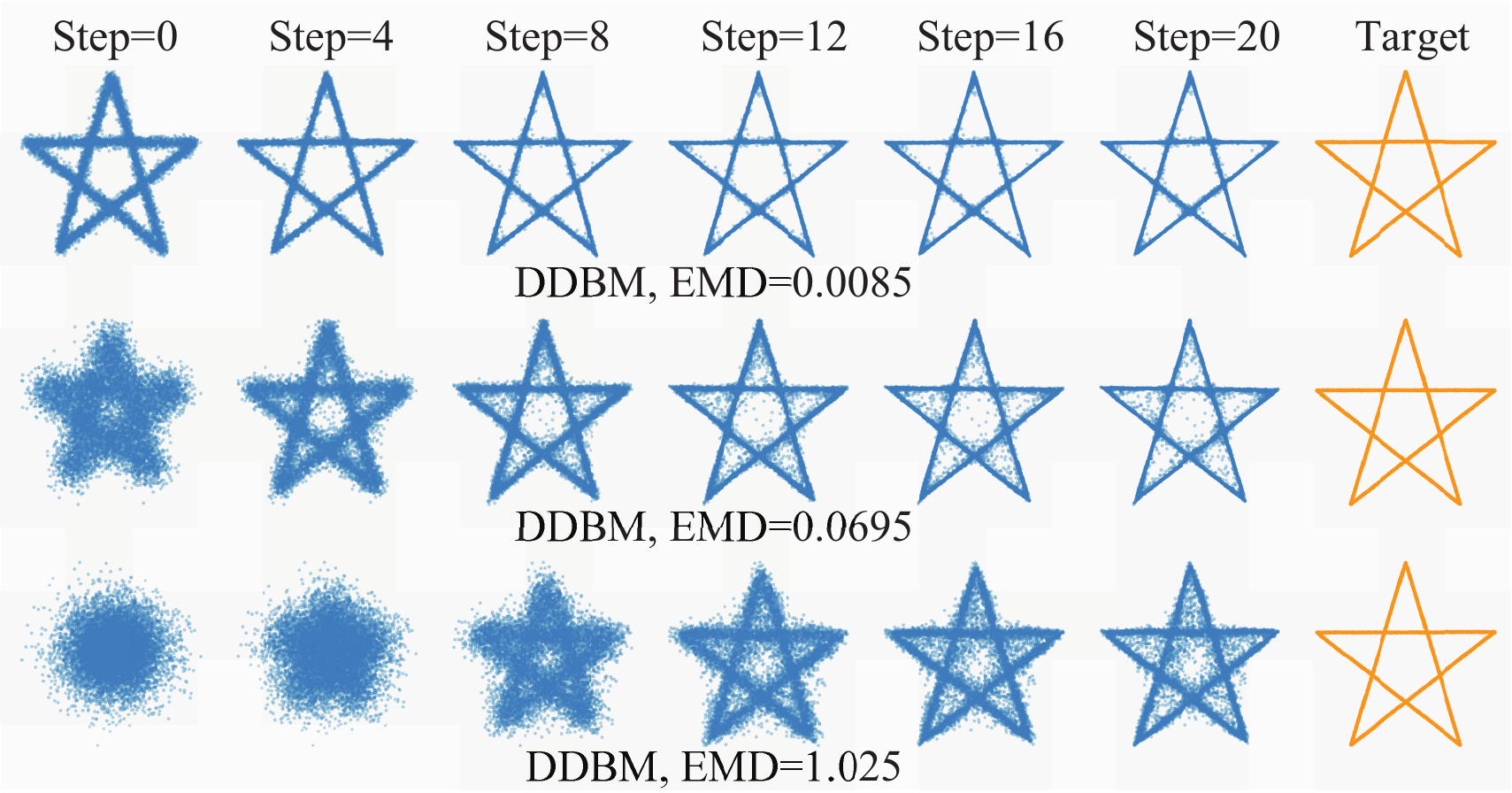}
    \caption{DDBM trained on 2D synthetic data. The leftmost side represents the source distribution, and the rightmost side represents the target distribution. The distance between the source distribution and the target distribution increases from top to bottom. As the EMD increases, the required approximation steps increase while model performance progressively declines.}
    \label{fig:toy}
\end{figure}

\textbf{Prior Action Generation}: 
As previously discussed, a source distribution that closely approximates the target distribution results in improved policy performance, underscoring the importance of an effective prior policy. A toy model is presented in Fig. \ref{fig:toy}.
For the sake of comparison, we propose the following three methods for generating the prior action:
\begin{itemize}
\item[$\bullet$] Gaussian Prior: 
As an uninformative prior, we select Gaussian white noise, consistent with that used in standard diffusion, as the source distribution, serving as the initial state in the denoising process. This choice facilitates direct comparison with methods based on the original diffusion approach.
\item[$\bullet$] Rule-based Prior:
We employ a simple, fully connected layer to map the contextual vector $\boldsymbol{c}_t$ to the action space, enabling the extraction of relevant low-dimensional features for action generation. The prior action is preset as a parabolic path passing through the current position, determined based on the predicted path length $d$ and movement behavior classification (straight, left turn, right turn, U-turn to the right, and U-turn to the left). Detailed information can be found in the Appendix.

\item[$\bullet$] Learning-based Prior:
This method involves generating prior actions by utilizing expert data and learning models. We use a lightweight conditional variational autoencoder (CVAE) as our learning model to achieve a lightweight design while maintaining effective generation. The observations \(\mathcal{O}\) are fed into the CVAE to produce corresponding actions. We aim for the model to effectively capture the significant correlations between observations and actions, thereby enabling a more efficient acquisition of the source distribution with reduced computational costs.
\end{itemize}

These methods for prior action generation are readily adaptable to other tasks related to imitation learning. Through a comparative analysis of these methods, we seek to assess the potential advantages of the diffusion bridge.

\subsection{Traing Details}
\textbf{Denoising Network}: We utilize a 1-D temporal CNN, conditioned on the $\boldsymbol{c}_t$ derived from observations, to iteratively denoise with the prior action $\mathbf{a}_T$ serving as the initial input. The network is modulated by $\boldsymbol{c}_t$ through Feature-wise Linear Modulation (FiLM) \cite{perez2018film} and performs $k$ iterations of denoising, following an approach similar to that of \cite{chi2023diffusion}.

\textbf{Loss Function}:
The loss function in our architecture comprises the diffusion bridge loss $\mathcal{L}_b$, the prior action generation loss $\mathcal{L}_p$, and a temporal distance loss $\mathcal{L}_d$ designed for high-level planning (following \cite{sridhar2024nomad}).

The diffusion bridge loss $\mathcal{L}_b$ is calculated by a weighted Mean Square Error (MSE) loss function:
\begin{equation}
\mathcal{L}_b = \mathbb{E}_{\mathbf{a}_t, \mathbf{a}_0, \mathbf{a}_T, t}\left[w(t)\|\mathbf{a}_{0}-D_{\theta}(\mathbf{a}_t, t, \mathbf{a}_T)\|^2\right],
\end{equation}

The prior action generation loss $\mathcal{L}_p$ is categorized into rule-based and learning-based approaches. The rule-based approach uses a fully connected layer to generate a coarse action. It is accompanied by a classification task that classifies the action based on its movement direction (details in the Appendix). Thus, the loss function consists of an action prediction loss computed by Mean Squared Error (MSE) and a classification loss based on cross-entropy:
\begin{equation}
\mathcal{L}_p^{rule}=-\lambda_c \sum p_c \log(\hat{p}_c)+\lambda_a\frac{1}{N} \sum_{i=1}^{N} \| \hat{\mathbf{a}}_i - \mathbf{a}_i \|^2,
\end{equation}
where $p_c$ represents the one-hot encoding of the ground truth class label, while $\hat{p}_c$ denotes the predicted probability. Additionally, $\mathbf{a}$ signifies the expert action, and $\hat{\mathbf{a}}$ represents the predicted action by the FC layer. The terms $\lambda_c$ and $\lambda_a$ denote the respective weights for the loss functions.

The learning-based approach employs the Conditional Variational Autoencoder (CVAE) model to efficiently generate actions in a lightweight manner, with the associated loss function defined as follows:
\begin{equation}
\mathcal{L}_p^{cvae}=\frac{1}{N} \sum_{i=1}^{N} \| \hat{\mathbf{a}}_i - \mathbf{a}_i \|^2+\text{KL}\left(\hat{\pi}(z|\mathcal{O},\mathbf{a}) \| \pi(z|\mathcal{O})\right),
\end{equation}
where $\hat{\pi}(\cdot)$ represents the posterior distribution of actions, while $\pi(\cdot)$ represents the prior.

The temporal distance loss $\mathcal{L}_d$ quantifies the interval count between the target image and the current image within a discretized timeline:
\begin{equation}
\mathcal{L}_d=\frac{1}{N} \sum_{i=1}^{N}\|f(\boldsymbol{c}_{t,i})-d_i\|^2,
\end{equation}
where $f(\boldsymbol{c}_{t,i})$ represents the prediction of temporal distance, and $d_i$ represents the ground truth.

In summary, the loss function is defined as:
\begin{equation}
\mathcal{L}=\lambda_b\mathcal{L}_b+\lambda_p\mathcal{L}_p+\lambda_d\mathcal{L}_d,
\end{equation}
where $\lambda_b$, $\lambda_p$ and $\lambda_d$ denote the respective weights for the loss functions.
\section{Experiments}
\label{sec:experiment}

\begin{table*}[ht!]
\centering
\caption{Quantitative comparison between the proposed method with baselines in simulation environments}
\begin{tabular}{cccccccc}
\hline
\multirow{2}{*}{\textbf{Scene}} & \multirow{2}{*}{\textbf{Method}} & \multicolumn{3}{c}{\textbf{Basic Task}} & \multicolumn{3}{c}{\textbf{Adaptation Task}} \\ \cline{3-8} 
& & \textbf{Length (m)} & \textbf{Collision} & \textbf{Success} & \textbf{Length (m)} & \textbf{Collision} & \textbf{Success} \\ \hline

\multirow{5}{*}{\begin{tabular}[c]{@{}c@{}}\textbf{Indoor}\\ (2D-3D-S)\end{tabular}}  
& ViNT~\cite{shah2023vint} & 151.1 $\pm$ 31.590 & 1.02 & 68\% & 19.9 $\pm$ 0.314 & 1.58 & 28\% \\ 
& NoMaD~\cite{sridhar2024nomad} & 147.9 $\pm$ 26.475 & 0.74 & 86\% & 20.3 $\pm$ 0.243 & 1.32 & 32\% \\ 
& Gaussian Prior & 148.7 $\pm$ 28.538 & 0.72 & 82\% & 19.8 $\pm$ 0.317 & 0.98 & 64\% \\ 
& Rule-based Prior & 163.57 $\pm$ 48.658 & 2.51 & 62\% & \textbf{19.0} $\pm$ 1.39 & 0.84 & 44\% \\ 
& Learning-based Prior & \textbf{143.7} $\pm$ \textbf{25.175} & \textbf{0.61} & \textbf{92\%} & 19.3 $\pm$ \textbf{0.34} & \textbf{0.41} & \textbf{88\%} \\ \hline

\multirow{5}{*}{\begin{tabular}[c]{@{}c@{}}\textbf{Outdoor}\\ (Citysim)\end{tabular}} 
& ViNT~\cite{shah2023vint} & 254.8 $\pm$ 63.597 & 0.78 & 20\% & 67.5 $\pm$ 26.320 & 0.41 & 38\% \\ 
& NoMaD~\cite{sridhar2024nomad} & 248.6 $\pm$ 67.679 & 0.58 & 22\% & 68.3 $\pm$ 26.315 & 0.34 & 52\% \\ 
& Gaussian Prior & 244.8 $\pm$ 62.749 & 0.62 & 34\% & 68.9 $\pm$ 24.656 & 0.35 & 50\% \\ 
& Rule-based Prior & 276.1 $\pm$ 83.721 & 3.59 & 14\% & 71.8 $\pm$ 30.856 & 0.59 & 28\% \\ 
& Learning-based Prior & \textbf{237.7} $\pm$ \textbf{59.476} & \textbf{0.51} & \textbf{44\%} & \textbf{63.5} $\pm$ \textbf{20.659} & \textbf{0.30} & \textbf{64\%} \\ \hline
\label{table:quantitative}
\end{tabular}
\vspace{-15pt}
\end{table*}

In this section, we comprehensively evaluate our method through simulated and real-world experiments in indoor and outdoor environments. The following sections provide an overview of the experimental setup, experiment results, and ablation studies.
\subsection{Experimental Setup}
\textbf{Datasets}.
Our method and all baseline methods were trained on the same dataset to ensure a fair comparison. Following the approach in~\cite{sridhar2024nomad}, the training data includes samples collected from diverse environments and various robotic platforms, incorporating data from RECON \cite{shah2021rapid}, SCAND \cite{karnan2022scand}, GoStanford \cite{hirose2019deep}, and SACSoN \cite{hirose2023sacson}. The dataset consists of sequences of consecutive image frames along with their corresponding positional information, providing comprehensive scene diversity.

\textbf{Baselines}.
We selected two state-of-the-art (SOTA) image-based visual navigation algorithms, NoMaD \cite{sridhar2024nomad} and ViNT \cite{shah2023vint}, for comparison. To examine the effect of integrating diffusion bridges on policy diffusion, we chose NoMaD, the first approach to apply policy diffusion to visual navigation tasks, which utilizes a standard diffusion policy model \cite{chi2023diffusion}. Additionally, we included ViNT, a regression-based model incorporating CNN and self-attention mechanisms, to compare the performance of generative and regression models in visual navigation tasks.

\textbf{Metrics}.
We report three evaluation metrics in our experiments to comprehensively assess the performance of the diffusion bridge-based visual navigation method: \\
\textit{Length}: For paths that complete the task, we calculate the path lengths' mean and variance to assess the navigation paths' efficiency and consistency.\\
\textit{Collision}: The average number of collisions per trial, which serves as an indicator of navigation safety.\\
\textit{Success Rate}: The success rate under consistent conditions. A trial is considered a failure if the robot does not reach the target position, becomes stuck due to collisions, or exceeds the time limit.

\textbf{Implementation details}.
We trained the model for 30 epochs using a learning rate 0.0001 and a batch size 256 (224 for learning-based methods). The training process, conducted on a single NVIDIA RTX TITAN GPU, took approximately 30 hours. For DDBM, we employed a VE model with parameters \(\sigma_0 = \sigma_T = 0.5\) and a default sampling step count of \(k = 10\).

\textbf{Evaluation Settings}.
To validate the effectiveness of the proposed method, we conducted comparative experiments in a simulated environment, benchmarking against baseline methods and performing ablation studies. These experiments were conducted in both indoor and outdoor settings. Furthermore, we evaluated the models on a robotic platform to assess their practical performance. The model operates on Nvidia Jetson Orin AGX deployed on the robot with RGB-only input.

\begin{figure*}[t]
\begin{center}
\includegraphics[width=\linewidth]{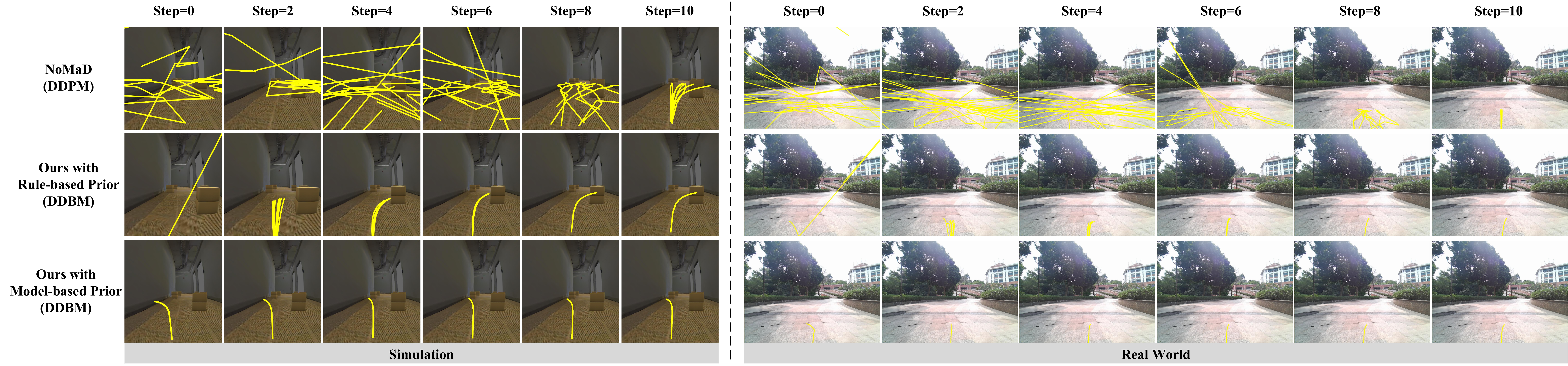}
\end{center}
\vspace{-10pt}
\caption{Comparison of navigation performance in simulation and real-world environments. NaviBridger with rule-based and model-based priors (DDBM) generates smoother, more stable paths compared to NoMaD (DDPM), demonstrating improved alignment with the target trajectory over denoising steps.}
\label{fig:exp2}
\vspace{-15pt}
\end{figure*}
\subsection{Results}
We conduct simulated experiments in two types of scenarios(Visualized in appendix): \textit{i)} Indoor (Stanford 2D-3D-S \cite{armeni2017joint}) \textit{ii)} Outdoor (Gazebo citysim \cite{koenig2004design}). 
In each scenario, we examine both basic and changed environments. The basic environment denotes a scenario where the test environment matches the configuration of the topological map, allowing us to evaluate long-distance navigation capabilities. The changed environment includes scenarios with obstacles in the test area that are not depicted in the topological map, aimed at testing the system's adaptability when the environment deviates from the mapped layout.

Fig. \ref{fig:exp2} includes visualizations of the denoising process in both simulated indoor environments and real-world environments. For detailed configurations of the robot in both simulations and the physical setup, please refer to the Appendix. It can be observed that the DDPM-based method begins denoising from Gaussian noise, with most of the denoising process dedicated to constraining noise unsuitable for the action space into the action space, and only in the final few steps does it constrain the actions required by the task. In contrast, our method starts from a favorable initial state, achieving stable actions after only a few denoising steps, such as 2 to 4. Even when starting with Gaussian noise, our model reaches a stable action output in fewer denoising steps. This demonstrates that a rich prior significantly enhances the convergence of policy diffusion. Moreover, under the DDBM framework, even when using uninformed Gaussian noise as the source distribution, the denoising approximate rate is substantially faster than that of the DDPM-based method.

The experimental results of the proposed method compared to the baseline in the simulation environment are presented in Table \ref{table:quantitative}. The learning-based prior method demonstrates optimal performance across various tasks and scenarios. The Gaussian prior method performs similarly to the DDPM-based NoMaD method, outperforming the baseline in the basic task within the outdoor scenario and the adaptation task within the indoor scenario. The rule-based method, on the other hand, only exceeds the baseline in the indoor adaptation task. These findings suggest that well-designed prior information can significantly improve the effectiveness of diffusion models, whereas poorly designed priors may impair model performance in specific scenarios. The rule-based method requires careful design of prior rules, which can yield significant improvements in some particular scenarios but generally lacks strong generalization across diverse conditions. When appropriate prior actions are unavailable, Gaussian noise is an acceptable initial distribution. A practical approach involves obtaining learned prior actions from a lightweight model and then generating actions through a diffusion bridge, offering a more versatile and robust solution. The learning-based method shows notable improvements in adaptation tasks and outdoor scenarios, indicating a solid capacity for generalization in complex and dynamic environments. In contrast, the DDPM-based method demonstrates comparatively weaker performance.

\subsection{Ablation Studies}

\textbf{The impact of denoising steps on model learning}. Table \ref{table:loss} shows the training results under different denoising step conditions, comparing the DDPM-based method and NaviBridger based on a Gaussian prior. We present the minimum mean squared error between the predicted action and the expert action on the training set and test sets of four datasets, along with the average ranking of each model across five items. Our method is closer to the expert trajectory in terms of overall data scale. Observing the trend, as the denoising steps decrease, the difference between the DDPM-based method's predicted trajectory and the expert trajectory increases significantly. In contrast, the difference between NaviBridger and a Gaussian prior remains relatively stable. This is also reflected in the average ranking; regardless of the number of denoising steps, Navibridger consistently achieves a better ranking, with minimal variation in average ranking as the denoising steps change. Conversely, the DDPM-based method shows generally lower rankings with more considerable fluctuations.

\begin{table}[h!]
\vspace{-5pt}
\centering
\setlength{\tabcolsep}{2pt}
\begin{tabular}{ccccccccc}
\toprule
\textbf{} & \multicolumn{4}{c}{\textbf{NoMaD (DDPM)}} & \multicolumn{4}{c}{\textbf{Gaussian Prior}} \\
\cmidrule(lr){2-5} \cmidrule(lr){6-9}
$k=$ & 10 & 7 & 4 & 1 & 10 & 7 & 4 & 1 \\
\midrule
\textbf{Train} & 1.38 & 1.27 & 1.14 & 2.16 & \textbf{0.27} &  0.28 & 0.30 & 0.30  \\
\textbf{Recon} & 1.19 & 1.10 & 0.97 & 1.43 & \textbf{0.88} & 0.91 & 0.89 & 0.98 \\
\textbf{Sacson} & 1.74 & 1.73 & 1.65 & 3.13 & 1.65 & \textbf{1.43} & 1.61 & 1.52 \\
\textbf{SCAND} & 0.72 & 0.76 & 0.71 & 1.17 & \textbf{0.59} & 0.69 & 0.63 & 0.61 \\
\textbf{Go Stanford} & 4.12 & 3.47 & 3.46 & 3.89 & 3.47 & 3.44 & \textbf{3.22} & 3.32 \\
\midrule
\textbf{Average Rank} & 7.0 & 6.0 & 4.4 & 7.8 & \textbf{2.4} & 2.6 & \textbf{2.4} & 2.8 \\
\bottomrule
\end{tabular}
\vspace{-5pt}
\caption{Comparison of the mean square error between DDBM and vanilla diffusion model in predicting actions and expert actions in training sets and different test sets.}
\label{table:loss}
\vspace{-10pt}
\end{table}

\begin{table}[h!]
\vspace{-10pt}
\centering
\setlength{\tabcolsep}{1pt}
\begin{tabular}{ccccc}
\toprule
\textbf{Prior} & \textbf{Method} & \textbf{Length (m)} & \textbf{Collision} & \textbf{Success} \\
\midrule
\multirow{2}{*}{\textbf{Gaussian}} & Source & -- & 2.12 & 0\%\\
 & Target &  19.8 $\pm$ 0.317 & 0.98 & 64\%\\
\multirow{2}{*}{\textbf{Rule-based}} & Source & 28.1 $\pm$ 5.942 & 1.77 & 8\%  \\
 & Target & 19.0 $\pm$ 1.390 & 0.84 & 44\%\\
\multirow{2}{*}{\textbf{Learning-based}} & Source & 24.8 $\pm$ 1.35 & 1.49 & 20\% \\
 & Target & 19.3 $\pm$ 0.340 & 0.41 & 88\%\\
\bottomrule
\end{tabular}
\vspace{-5pt}
\caption{Performance metrics of visual navigation tasks across different source priors: evaluation of policy effectiveness for source-target action mapping.}
\label{table:source}
\vspace{-10pt}
\end{table}

\textbf{The effect of source distribution and the role of diffusion bridge}. Table \ref{table:source} compares the navigation performance of the proposed method in an indoor adaptation task by examining prior actions sampled from different source distributions and the target actions obtained through denoising. As the prior generation method transitions from random to ordered and from heuristic-based to expert data-driven approaches, the success rate of navigation using only the prior actions shows some improvement, although it remains pretty low. After bridge denoising, the navigation success rate improves markedly. The experiments indicate that more informative prior actions contribute to better performance. At the same time, the diffusion bridge effectively transforms less effective prior actions into higher-quality target actions, thereby enhancing overall performance.
\begin{figure}[htbp]
    \centering
    \includegraphics[width=\linewidth]{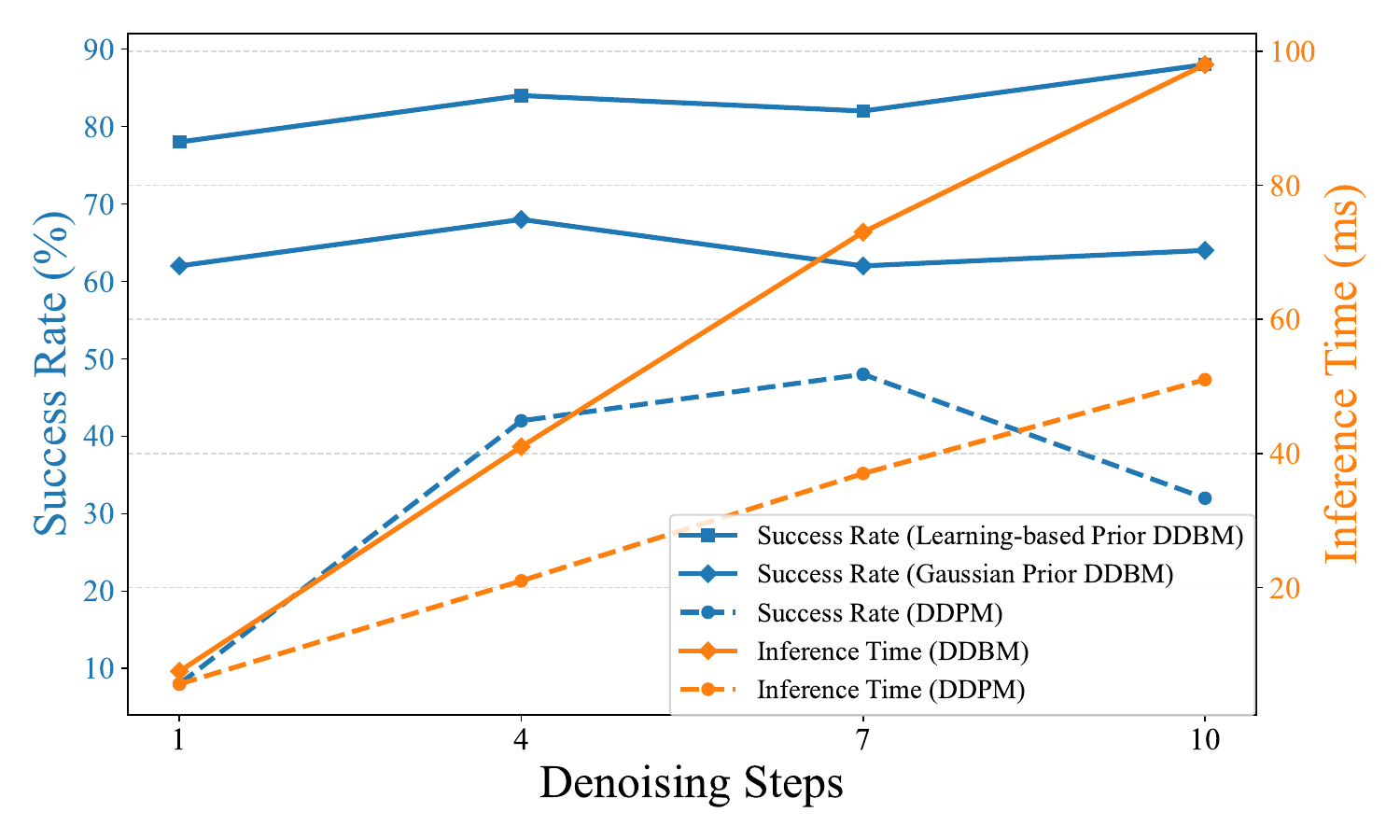}
    \vspace{-5pt}
    \caption{Success rate and inference time comparison across denoising steps. Learning-based and Gaussian priors in DDBM show higher success rates than DDPM, with learning-based priors achieving the best performance. DDBM also demonstrates faster inference times as denoising steps increase.}
    \label{fig:linechart}
    \vspace{-15pt}
\end{figure}

\textbf{Less denoising steps and the same performance.}
The line chart as shown in Fig \ref{fig:linechart} illustrates the relationship between denoising steps, success rate, and inference time for visual navigation under different prior settings—Learning-based Prior DDBM, Gaussian Prior DDBM, and DDPM. Success rate, represented on the left vertical axis, generally increases as denoising steps progress, with the Learning-based Prior DDBM consistently achieving the highest success rates, peaking around 90\% at 10 denoising steps. In contrast, the Gaussian Prior DDBM shows a moderate success rate that fluctuates slightly, while DDPM achieves a lower success rate overall. The inference time indicated on the right vertical axis increases linearly with the denoising steps, and the time required by DDBM is longer than that of DDPM under the same denoising steps. Learning-based Prior DDBM demonstrates a more favorable balance between high success rate and moderate inference time, suggesting it is more efficient and reliable for visual navigation tasks than the Gaussian Prior DDBM and DDPM.
\section{Conclusion}
\label{sec:conclusion}

This paper proposes a novel diffusion bridge-based framework for visual navigation. Our method advances policy generation by enabling translations across arbitrary distributions based on informative prior actions rather than the random Gaussian noise. Our theoretical analysis details fundamental principles, formulas, and the influence of source distributions on performance, along with an analysis of error lower bounds. Experimental results demonstrate a substantial increase in success rates and a reduction in collisions compared to baseline models in simulated environments. By exploring different strategies for source distribution generation and denoising steps, our method achieves greater robustness, more stability, and faster action generation than vanilla diffusion policy models, expanding the potential of policy diffusion in imitation learning.

However, although the results demonstrate a certain extent of generalization to dynamic scenarios, the scenario becomes more complex if high-speed moving objects are present, necessitating faster response or adaptive strategies.

\textbf{Acknowledgment}. This work is supported by the China National Key R\&D Program (2022YFB3903804).
{
    \small
    \bibliographystyle{ieeenat_fullname}
    \bibliography{main}
}

\clearpage
\maketitlesupplementary
\appendix

\section{Details of DDBM in Imitation Learning}
As shown in Section \ref{subsec:ddbmil}, the evolution of conditional probability of denoising diffusion bridfge models $q(\mathbf{a}_t|\mathbf{a}_T)$ has a time-reversed SDE of the form:
\begin{equation}
\begin{split}
d\mathbf{a}_t &= [f(\mathbf{a}_t, t) - g(t)^2 (s(\mathbf{a}_t, t, \mathbf{a}_T, T) \\
            &\quad - h(\mathbf{a}_t, t, \mathbf{a}_T, T))] dt + g(t) d\bar{\mathbf{w}}_t,
\end{split}
\end{equation}
with an associated probability flow ODE:
\begin{equation}
\label{eq:ode}
\begin{split}
d\mathbf{a}_t &= [f(\mathbf{a}_t, t) - g(t)^2 (\frac{1}{2}s(\mathbf{a}_t, t, \mathbf{a}_T, T) \\
            &\quad - h(\mathbf{a}_t, t, \mathbf{a}_T, T))] dt + g(t),
\end{split}
\end{equation}
on $t<T-\epsilon$ for any $\epsilon>0$, where $\mathbf{\hat{w}}_t$ denotes a Wiener process.

The derivation of $q(\mathbf{a}_t|\mathbf{a}_0, \mathbf{a}_T)$ is shown at Section \ref{subsec:ddbmil}. Clearly defining each variable in the formula is necessary to complete the computation process. Following the definition of \cite{zhou2024denoising}, the bridge processes generated by both VP and VE diffusion are in Table \ref{table:instantiations}.

As mentioned in Section \ref{subsec:ddbmil}, with the reparametrization of Elucidating Diffusion Models, score function $\mathbf{s}(\mathbf{a},t,\mathbf{a}_T,T)$ can be expressed as:

\begin{equation}
\label{eq:reparam1}
\begin{aligned}
  &\nabla_{\mathbf{a}_t} \text{log}q(\mathbf{a}_t|\mathbf{a}_T) \approx \mathbf{s}({D_\theta}, \mathbf{a}_t, t, \mathbf{a}_T, T) \\ &= \frac{\mathbf{a}_t - (\frac{SNR_T}{SNR_t}\frac{\alpha_t}{\alpha_T}\mathbf{a}_T + \alpha_t D_{\theta}(\mathbf{a}_t,t, \mathbf{a}_T)(1-\frac{SNR_T}{SNR_t}))}{\sigma^2_t(1-\frac{SNR_T}{SNR_t})},
\end{aligned}
\end{equation}

We following DDBM \cite{zhou2024denoising} to define the scaling functions and weighting function $w(t)$:

\begin{equation}
c_{\text{in}}(t) = \frac{1}{\sqrt{a_t^2\sigma_T^2 + b_t^2\sigma_0^2 + 2a_tb_t\sigma_0T + c_t}}
\end{equation}
\begin{equation}
c_{\text{skip}}(t) = \left(b_t\sigma_0^2 + a_t\sigma_0T\right) \cdot c_{\text{in}}^2(t)
\end{equation}
\begin{equation}
c_{\text{out}}(t) = \sqrt{a_t^2\left(\sigma_T^2\sigma_0^2 - \sigma_0T^2\right) + \sigma_0^2c_t \cdot c_{\text{in}}(t)}
\end{equation}
\begin{equation}
w(t) = \frac{1}{c_{\text{out}}(t)^2}
\end{equation}
\begin{equation}
c_{\text{noise}}(t) = \frac{1}{4} \log(t)
\end{equation}
where $a_t = \frac{\alpha_t}{\alpha_T} \cdot \frac{\text{SNR}_T}{\text{SNR}_t}, \quad b_t = \alpha_t \left(1 - \frac{\text{SNR}_T}{\text{SNR}_t}\right), \quad c_t = \sigma_t^2 \left(1 - \frac{\text{SNR}_T}{\text{SNR}_t}\right)$.

Following EDM \cite{karras2022elucidating}, the model output can be parameterized as:
\begin{equation}
D_{\theta}(\mathbf{a}_t, t) = c_{\text{skip}}(t)\mathbf{a}_t + c_{\text{out}}(t)F_{\theta}\big(c_{\text{in}}(t)\mathbf{a}_t, c_{\text{noise}}(t)\big),
\end{equation}
where $F_{\theta}$ is a neural network with parameter $\theta$ that predicts $x_0$.

\begin{table*}[htp]
    \centering
    \begin{tabular}{cccccc}
        \toprule
        \textbf{ } & \textbf{f$(\mathbf{a}_t, t)$} & \textbf{$g^2(t)$} & \textbf{$p(\mathbf{a}_t \mid \mathbf{a}_0)$} & \textbf{SNR$_t$} & \textbf{$\nabla_{\mathbf{a}_t} \log p(\mathbf{a}_T \mid \mathbf{a}_t)$} \\
        \midrule
        VP & $\frac{d \log \alpha_t}{dt} \mathbf{a}_t$ & $\frac{d}{dt} \sigma_t^2 - 2 \frac{d \log \alpha_t}{dt} \sigma_t^2$ & $\mathcal{N}(\alpha_t \mathbf{a}_0, \sigma_t^2 \mathbf{I})$ & $\frac{\alpha_t^2}{\sigma_t^2}$ & $\frac{\left(\frac{\alpha_t}{\alpha_T} \mathbf{a}_T - \mathbf{a}_t\right)}{\sigma_t^2 \left(\text{SNR}_t / \text{SNR}_T - 1\right)}$ \\
        VE & $0$ & $\frac{d}{dt} \sigma_t^2$ & $\mathcal{N}(\mathbf{a}_0, \sigma_t^2 \mathbf{I})$ & $\frac{1}{\sigma_t^2}$ & $\frac{\mathbf{a}_T - \mathbf{a}_t}{\sigma_T^2 - \sigma_t^2}$ \\
        \bottomrule
    \end{tabular}
    \caption{VP and VE instantiations of diffusion bridges.}
    \label{table:instantiations}
\end{table*}

\section{Rule-based Prior Design and Derivation}
\subsection{Rule-based Prior Design}
The rule-based prior design employed in this work aims to generate an initial action that reflects plausible movement patterns based on the contextual features $\boldsymbol{c}_t$. To achieve this, the action space is divided into five distinct decision categories: moving straight, turning left, turning right, making a U-turn to the left, and making a U-turn to the right. This partitioning simplifies the prior generation process while ensuring comprehensive coverage of potential movement behaviors.

We utilize a fully connected layer to map the contextual vector $\boldsymbol{c}_t$ into a low-dimensional representation of action-relevant features to establish the prior. These features explicitly include the decision category and the predicted path length $d$. This mapping effectively encapsulates the essential elements required for generating an initial action, balancing the complexity of motion planning with computational efficiency.

Based on the decision category and path length derived from the contextual features, the initial direction and length of the path are determined. To introduce variability and prevent overly deterministic priors, noise is added to the angular direction and path length. This noise ensures diversity in the generated prior actions, enabling the model to explore a broader range of potential actions while maintaining adherence to the underlying behavioral patterns.

The initial action is then constructed following a parabolic path definition. The parabolic path is formulated such that it passes through the starting point and the determined endpoint. By leveraging the properties of parabolic curves, the path naturally accommodates smooth transitions and realistic motion patterns. We sample a fixed number of evenly spaced points along the parabola to represent this curve, providing a discrete sequence that constitutes the initial action.

This rule-based approach offers a robust and interpretable framework for generating prior actions, serving as a foundation for subsequent optimization and refinement. Integrating decision categories, noise, and parabolic curve modeling ensures that the priors are both expressive and adaptable, facilitating effective downstream action generation. Further details on the mathematical formulation of the parabolic path and noise modeling can be found in Section \ref{subsec:parabolic}.
\subsection{Parabolic Constraint Condition Derivation}
\label{subsec:parabolic}
The desired path is modeled using a family of parabolas that meet the following constraints: the trajectory passes through the robot's origin and a specified endpoint. The initial direction aligns with the robot's forward orientation, transitioning smoothly to the endpoint in a parabolic arc. The parabola's vertex lies between the starting point and the endpoint, with its opening directed toward the robot's rear.

The parabolic constraint condition is derived based on the standard and vertex forms of a parabola, ensuring the curve passes through two fixed points \((x_1, y_1)\) and \((x_2, y_2)\), with control over its shape and orientation. The standard form of a parabolic equation is \( y = ax^2 + bx + c \), where \(a\), \(b\), and \(c\) are parameters. To uniquely define the parabola, three conditions are required. By incorporating the axis of symmetry \(h\) as an additional control condition, a family of parabolas can be generated through the fixed points.

Alternatively, the vertex form, \(y = a(x - h)^2 + k\), provides a parameterization where \(h\) represents the axis of symmetry, \(k\) denotes the vertical coordinate of the vertex, and \(a\) determines the parabola's width and direction. Substituting the fixed points into the vertex form equations:

\begin{equation}
y_1 = a(x_1 - h)^2 + k \quad \text{and} \quad y_2 = a(x_2 - h)^2 + k,
\end{equation}

the parameter \(a\) can be derived by eliminating \(k\):

\begin{equation}
a = \frac{y_2 - y_1}{(x_2 - x_1)(x_2 + x_1 - 2h)}.
\end{equation}

Substituting \(a\) back, the expression for \(k\) becomes:

\begin{equation}
k = y_1 - a(x_1 - h)^2.
\end{equation}

This formulation results in a family of parabolas passing through the fixed points, given as:

\begin{equation}
    \begin{split}
y &= \frac{y_2 - y_1}{(x_2 - x_1)(x_2 + x_1 - 2h)}(x - h)^2 + y_1\\
&- \frac{y_2 - y_1}{(x_2 - x_1)(x_2 + x_1 - 2h)}(x_1 - h)^2
    \end{split}
\end{equation}

To ensure the parabola opens downward, the condition \(a < 0\) must hold. Since \(a = \frac{y_2 - y_1}{(x_2 - x_1)(x_2 + x_1 - 2h)}\), the numerator and denominator must have opposite signs. The numerator’s sign depends on the relative magnitudes of \(y_2\) and \(y_1\), while the denominator’s sign is determined by \((x_2 - x_1)(x_2 + x_1 - 2h)\).

For cases where \(y_2 > y_1\), the numerator is positive, requiring the denominator to be negative. If \(x_2 > x_1\), this implies \(h > \frac{x_1 + x_2}{2}\). Conversely, if \(x_2 < x_1\), \(h < \frac{x_1 + x_2}{2}\). Similarly, when \(y_2 < y_1\), the numerator is negative, necessitating a positive denominator. For \(x_2 > x_1\), this requires \(h < \frac{x_1 + x_2}{2}\), and for \(x_2 < x_1\), \(h > \frac{x_1 + x_2}{2}\). These relationships enable precise control over the parabola’s orientation.

The position of the vertex relative to the fixed points is governed by the axis of symmetry \(h\). To place the vertex between \(x_1\) and \(x_2\), \(x_1 < h < x_2\) or \(x_2 < h < x_1\) must hold. Conversely, for the vertex to lie outside this range, \(h < \min(x_1, x_2)\) or \(h > \max(x_1, x_2)\) is required. Adjusting \(h\) accordingly provides flexibility in the parabola’s configuration while satisfying the desired constraint conditions.

\subsection{Noise Modeling Description}
In this framework, the standard deviation of Gaussian noise is adaptively adjusted based on prediction confidence. Higher confidence leads to lower noise, ensuring precision, while lower confidence increases noise to encourage exploration. The relationship is defined by:

\begin{equation}
\sigma = \text{min\_std} + (\text{max\_std} - \text{min\_std}) \cdot (1 - \text{confidence}),
\end{equation}
where \( \text{confidence} \) ranges from 0 to 1. This linear interpolation maps confidence to a noise level between predefined minimum \( \text{min\_std} \) and maximum \( \text{max\_std} \) values. 

The adaptive noise is applied to both angular direction and path length. For angular direction, the noise is centered on the midpoint of the decision interval, while for path length, it is centered on the predicted value, with variability scaled by the corresponding confidence.

\begin{figure*}[htbp]
    \centering
    \includegraphics[height=11.5cm,angle=90]{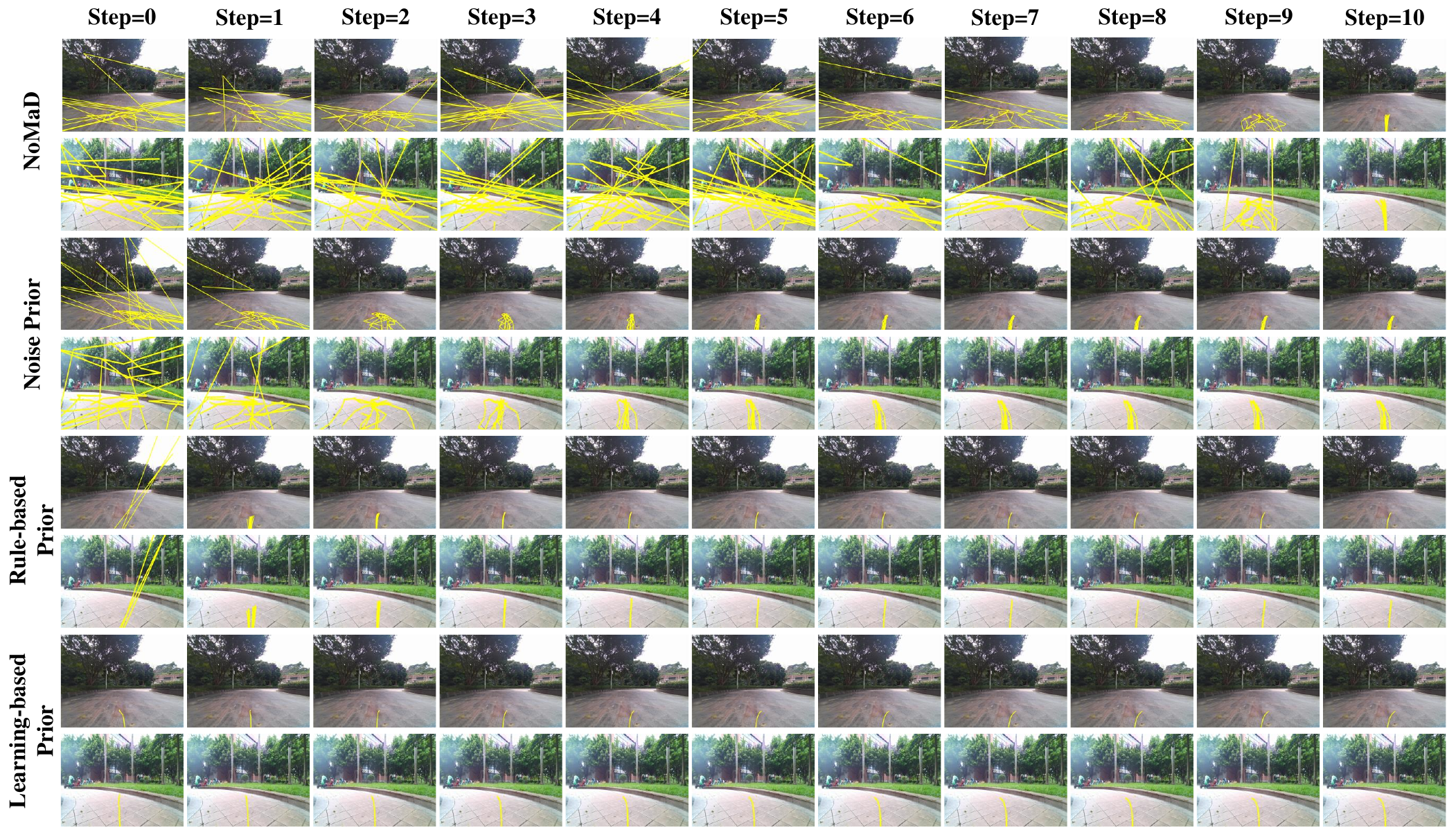}
    \caption{The figure visualizes feature correspondences over ten steps for four methods: NoMaD, Noise Prior, Rule-Based Prior, and Learning-Based Prior, in the context of on-device RGB observations. From top to bottom, the rows represent the progression of steps (Step 0 to Step 10). At the same time, from left to right, the columns correspond to the four method types—NoMaD, Noise Prior, Rule-Based Prior, and Learning-Based Prior—showcasing the impact of each prior on the feature-matching process across the sequence.}
    \label{afig:1}
\end{figure*}

\begin{figure*}[thbp]
    \centering
    \includegraphics[width=\linewidth]{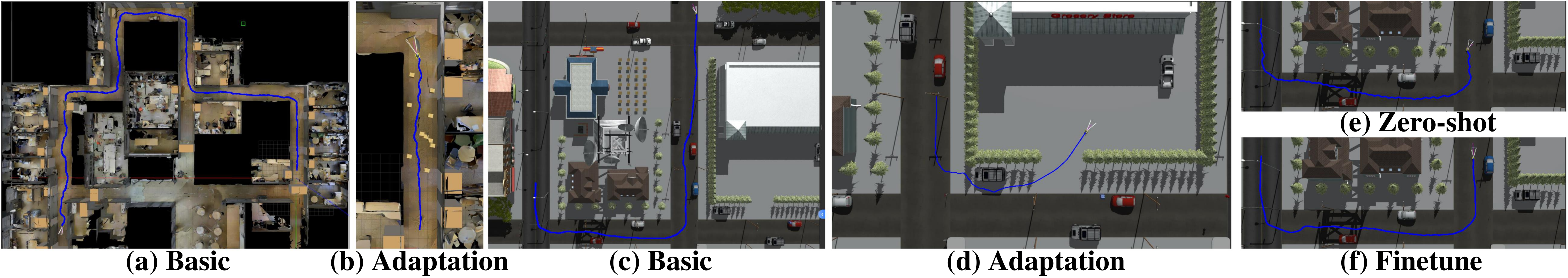}
    \caption{Simulation Tasks and Zero-Shot vs. Fine-Tuned Model}
    \label{afig:3}
    \vspace{-10pt}
\end{figure*}

\section{Detailed Analysis of Error Bound}

The key goal is to bound the mean squared error (MSE), represented as $\mathbb{E}_{\mathbf{a}_t, \mathbf{a}_0} \left[ \| \mathbf{a}_t - \mathbf{a}_0 \|^2 \right]$. 

This error is driven by the difference between the source distribution \( \pi_s(\mathbf{a}_T) \) and the target distribution \( \pi(\mathbf{a}_0) \). The DDBM framework is designed to reduce this discrepancy through a diffusion process and a drift term in its probability flow ODE.

The initial difference between the source and target distributions is fundamental to the bound. As shown in Section \ref{subsec:priordesign}, we represent this discrepancy using the Kullback-Leibler (KL) divergence:

\begin{equation}
D_{\text{KL}}\left(\pi_{s}(\mathbf{a}_T) \| \pi(\mathbf{a}_0)\right) = \int \pi_{s}(\mathbf{a}_T) \log \frac{\pi_{s}(\mathbf{a}_T)}{\pi(\mathbf{a}_0)} d\mathbf{a}_T.
\end{equation}

This KL divergence measures how much information is "lost" if we approximate \( \pi(\mathbf{a}_0) \) by \( \pi_s(\mathbf{a}_T) \). A smaller divergence indicates that the source and target distributions are close, implying that a smaller amount of drift and noise adjustment will be necessary for the DDBM to match the target.

The evolution of samples in a DDBM is controlled by a \textbf{probability flow ODE} that includes a \textbf{drift term} and a \textbf{noise term}. The ODE has the form which is equal to Equation \ref{eq:ode}:

\begin{equation}
    \frac{d\mathbf{a}_t}{dt}\! =\! f(\mathbf{a}_t, t) - g^2(t)\! \left(\!\frac{1}{2} \nabla_{\mathbf{a}_t}\! \log \pi(\mathbf{a}_t | \mathbf{a}_0)\! -\! \nabla_{\mathbf{a}_t}\! \log \pi(\mathbf{a}_0 | \mathbf{a}_t)\!\right)\!, 
\end{equation}
where \( f(\mathbf{a}_t, t) \) is a deterministic part of the drift that ensures gradual movement from the source to the target distribution. \( g^2(t) \) controls the level of noise at time \( t \), modulated by the signal-to-noise ratio (SNR).

The term \( \frac{1}{2} \nabla_{\mathbf{a}_t} \log \pi(\mathbf{a}_t | \mathbf{a}_0) - \nabla_{\mathbf{a}_t} \log \pi(\mathbf{a}_0 | \mathbf{a}_t) \) is a correction factor that adjusts the position of \( \mathbf{a}_t \) to bring it closer to \( \mathbf{a}_0 \). This adjustment term is critical for reducing error in the final output and directly depends on the initial discrepancy \( D_{\text{KL}}(\pi_s(\mathbf{a}_T) \| \pi(\mathbf{a}_0)) \).

The initial discrepancy contributes a fundamental limit to how close the source and target can be through the evolution process. The noise \( g(t) \) and the signal-to-noise ratio \( \text{SNR}_t \) at each time step both affect how much control we have over the diffusion process. Higher \( g(t) \) implies more noise, making it harder for the model to accurately adjust the path toward the target. Higher SNR (meaning a larger signal relative to noise) allows for more precise alignment with \( \mathbf{a}_0 \) as the process evolves.

To incorporate these elements into the error bound, we arrive at the following formula: 
\begin{equation}
\mathbb{E}_{\mathbf{a}_t, \mathbf{a}_0} \left[ \| \mathbf{a}_t - \mathbf{a}_0 \|^2 \right]
\leq \frac{1}{2} \cdot \frac{g^2(t)}{\text{SNR}_t} \cdot D_{\text{KL}}(\pi_s(\mathbf{a}_T) \| \pi(\mathbf{a}_0)), 
\end{equation}
where \( \frac{1}{2} \) arises from the specific form of the adjustment term in the probability flow ODE. \( \frac{g^2(t)}{\text{SNR}_t} \) shows how the noise and signal-to-noise ratio affect the error bound. Because it is invariant under a fixed schedule, it can be expressed as a constant $C$.

\section{Detailed Experimental Settings and Results}

\begin{figure*}[htbp!]
    \centering
    \includegraphics[width=0.9\linewidth]{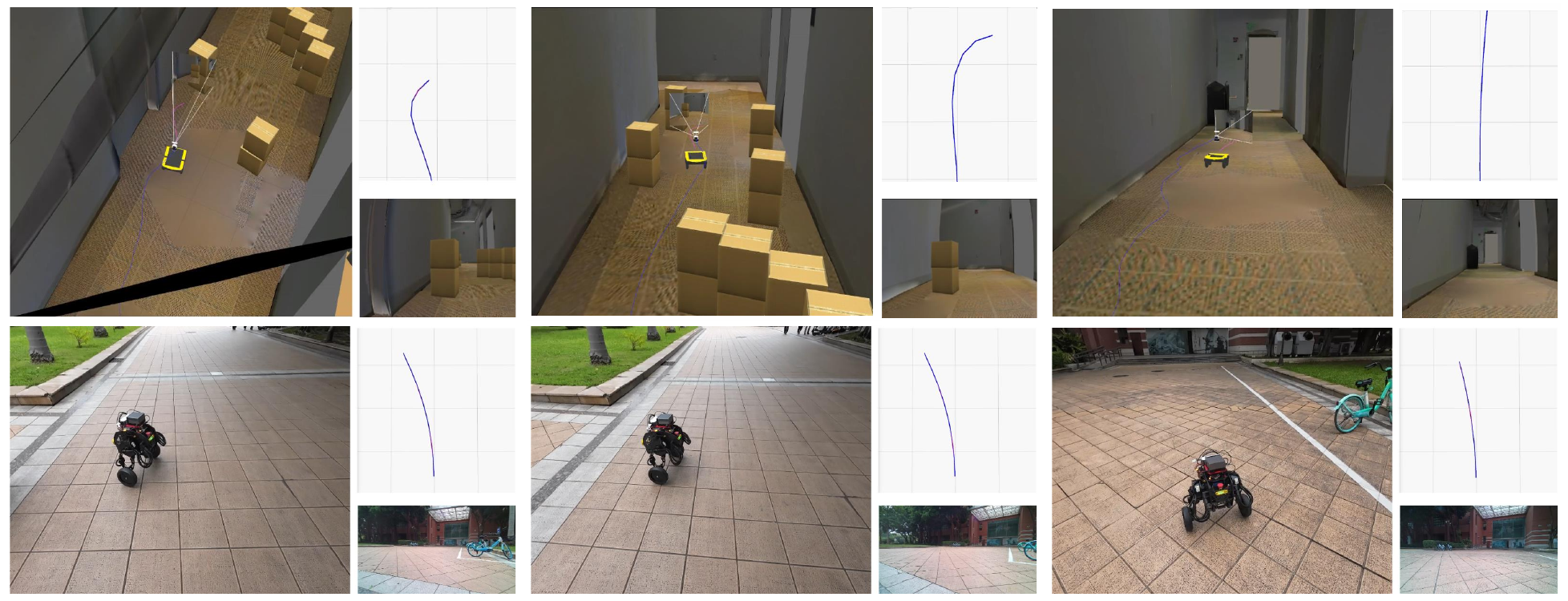}
    \caption{The figure presents navigation trajectories from experiments in both simulated and real-world environments. The top row shows obstacle navigation in a simulated corridor with visual observations and trajectory curves, while the bottom row displays outdoor navigation on tiled pathways with real camera observations. The columns highlight different test scenarios, illustrating the robot's ability to adapt and maintain consistent navigation strategies across varied environments.}
    \label{afig:2}
    \vspace{-10pt}
\end{figure*}

The experimental setup in the simulation environment is illustrated in Fig. \ref{afig:3}, where (a) and (c) represent indoor and outdoor base tasks, while (b) and (d) show their adaptation counterparts. For adaptation tasks, significant environmental differences exist between the target image collection phase and navigation execution phase: randomly placed boxes (b) and vehicles (d) along paths are absent in target images but present during algorithm operation, validating environmental adaptability. Subfigures (e)-(f) evaluate zero-shot generalization capabilities. Given the substantial domain gap between our publicly available real-world training dataset and simulation/test-bed environments (particularly in visual appearance), (e) demonstrates pure zero-shot performance without simulation data, while (f) shows improved results after fine-tuning with limited simulation trajectory data unrelated to target tasks. Experimental results confirm the algorithm's strong zero-shot transfer capability, with notable performance gains achieved through minimal fine-tuning.

\begin{figure}[htbp]
    \centering
    \includegraphics[width=0.9\linewidth]{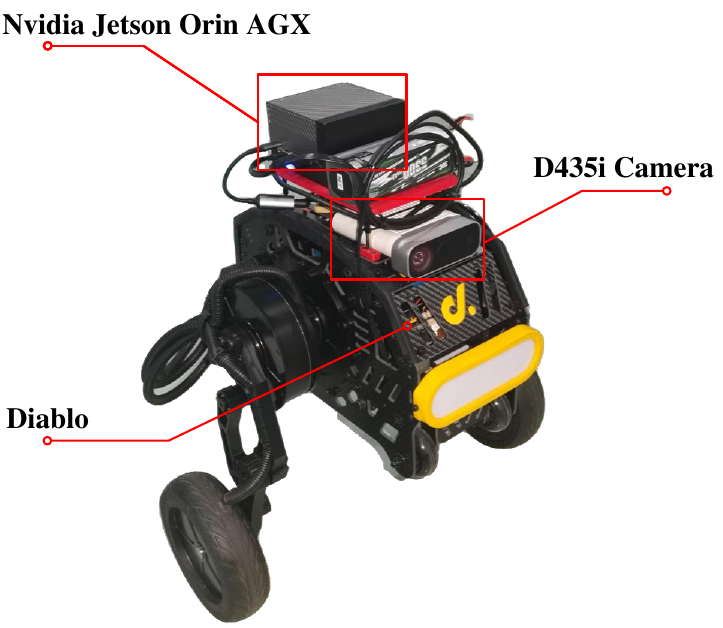}
    \caption{The wheeled-legged robot named Diablo, equipped with an Intel RealSense D435i camera and an Nvidia Jetson Orin AGX, was utilized for on-device experiments relying solely on RGB observations.}

    \label{afig:robot}
\end{figure}

The metrics in Table \ref{table:source} and Fig. \ref{fig:linechart} are derived from adaptation tasks in indoor scenarios, while the real-world results in Fig. \ref{fig:exp2} are based on on-device experiments. All the simulation experiments were deployed on a Nvidia RTX2080Ti GPU, utilizing a Jackal move robot. The on-device experiments utilized the wheeled-legged robot Diablo, equipped with an Intel Realsense D435i and relying solely on RGB observations as shown in Fig. \ref{afig:robot}. As mentioned in the main text, the model was deployed on a Nvidia Jetson Orin AGX. The complete navigation process relies on a high-level planner based on a topological map. Sub-goals on the topological map are used as target image inputs for the network, guiding the policy toward the destination and enabling long-distance navigation.

Fig. \ref{afig:1} illustrates the complete denoising process for NoMaD and NaviBridger with different prior actions based on DDPM. It can be observed that NaviBridger achieves higher denoising efficiency, requiring only a few steps to complete the process, whereas NoMaD based on DDPM needs more steps. Moreover, as the prior action becomes closer to the target action, the difference between the source and target distributions decreases, reducing the difficulty of the denoising network and leading to better action results.

Fig. \ref{afig:2} illustrates the results of navigation experiments using the learning-based prior approach in both simulated and real-world environments, highlighting its zero-shot generalization capabilities. The top row demonstrates the robot navigating a simulated corridor with static obstacles, where smooth and collision-free trajectories are achieved without prior data collection from similar test environments. The rendered visualizations and robot perspectives further emphasize the system's ability to adapt seamlessly to structured indoor scenarios. The bottom row presents the robot navigating outdoor real-world environments on tiled pathways, showcasing consistent and effective trajectories despite differences in terrain and environmental context. The learning-based prior method excels in adaptability, exhibiting robust performance across diverse scenarios and robot configurations. This zero-shot capability demonstrates its ability to generalize navigation strategies without needing domain-specific data or adjustments for different robots. The video is available in the supplementary materials.

\end{document}